# LEARNet: Dynamic Imaging Network for Micro Expression Recognition

Monu Verma, Santosh Kumar Vipparthi, Girdhari Singh, Subrahmanyam Murala

*Abstract—* Unlike prevalent facial expressions, micro expressions have subtle, involuntary muscle movements which are short-lived in nature. These minute muscle movements reflect true emotions of a person. Due to the short duration and low intensity, these micro-expressions are very difficult to perceive and interpret correctly. In this paper, we propose the dynamic representation of micro-expressions to preserve facial movement information of a video in a single frame. We also propose a Lateral Accretive Hybrid Network (LEARNet) to capture micro-level features of an expression in the facial region. The LEARNet refines the salient expression features in accretive manner by incorporating accretion layers (AL) in the network. The response of the AL holds the hybrid feature maps generated by prior laterally connected convolution layers. Moreover, LEARNet architecture incorporates the cross decoupled relationship between convolution layers which helps in preserving the tiny but influential facial muscle change information. The visual responses of the proposed LEARNet depict the effectiveness of the system by preserving both high- and micro-level edge features of facial expression. The effectiveness of the proposed LEARNet is evaluated on four benchmark datasets: CASME-I, CASME-II, CAS(ME)^2 and SMIC. The experimental results after investigation show a significant improvement of 4.03%, 1.90%, 1.79% and 2.82% as compared with ResNet on CASME-I, CASME-II, CAS(ME)^2 and SMIC datasets respectively.

## I. INTRODUCTION

Facial expressions can provide a rich source of affective information in daily life social communication. Usually, facial expressions (macro) last for about 4 to 5 seconds and are easy for humans to perceive. However, certain emotions manifest themselves as micro-expressions, which are very rapid (last only for 1/3 to 1/25 seconds) and involuntary in nature. These micro-expressions give us a glimpse or insight of the true emotions/ feelings of a person even if he/she is trying to hide the actual emotions through false macro-expressions. Ekman [1] identified various deceptive expressions in an interview video of a depressed patient, who attempted to commit suicide. He analyzed that the patient represses his/her intensive sadness in smile within 1/12 seconds. Though, those expressions were spotted in a few frames of video, captured through standard 25 fps devices but provided enough clues to recognize the true sentiments of a person. Since, the micro expressions occur in fraction of a second and have very low intensity, it is very hard to detect them through human endeavors. Only professionally trained persons can spot and identify these expressions. Even with professional training, only 47% recognition accuracy has been reported in the literature [2]. Automatic micro expression recognition is yielding more attention due to its wide range of applications in various fields: police interrogation, clinical diagnosis, depression analysis, lie detection, business negotiation, teaching assistance, law enforcement etc.

Ekman [3] categorized emotions into six universal classes as: anger, disgust, fear, happy, sad and surprise. Moreover, Ekman et al. [4-5] developed facial action coding system (FACS) and micro-expression training tool (METT) to standardize the automated facial expression recognition (FER) systems. They divided the facial image into small units based on muscle movements and categorized them as action units (AUs). An automated micro expression recognition (MER) system mainly consists of three steps: face detection and preprocessing, spatio-temporal feature extraction and expression classification. In the first step, facial region of interest is extracted by subtracting background noise. In the second step, spatio-temporal features are extracted from the preprocessed images. In spatio-temporal features, spatio implies facial appearance at a particular time instance and temporal represents the momentary changes arising between the frames or frame sequences. Spatio-temporal feature extraction method has an essential role in any MER system. Substandard features would lead to a higher number of false positives and false negatives thereby degrading the performance of the MER system even with the best of the classifiers. Thus, designing a robust feature descriptor is essential for the MER system. In the literature, many descriptors have been proposed to extract the spatio-temporal features of the micro-expression such as LBP-TOP [6], TIM [7], DTSA [8], TICS [9-11] etc. In the last step, the extracted pertinent features are forwarded to a classifier, which identifies the feature's characteristics to make appropriate classification decision.

Quite recently, with the advancement in smart technologies such as machine learning, artificial intelligence and robust hardware design (graphical processing unit (GPU)), deep learning has become the most effective learning technique in

Monu Verma, Santosh Kumar Vipparthi and Girdhari Singh are with Department of Computer Science and Engineering, Malaviya National Institute of Technology, Jaipur, India (Email: monuverma.cv@gmail.com; skvipparthi@mnit.ac.in; gsingh.cse@mnit.ac.in )

Subrahmanyam Murala is with Department of Electrical Engineering, Indian Institute of Technology Ropar, Roopnagar, India (Email: subbumurala@iitrpr.ac.in ) areas like object detection, face recognition, facial verification, facial expression recognition etc. Many deep learning models like VGG-16, VGG-19 [12], AlexNet [13], Google Net [14], ResNet [15] etc. have been introduced in the literature that have achieved and improved performance in the above-mentioned fields.

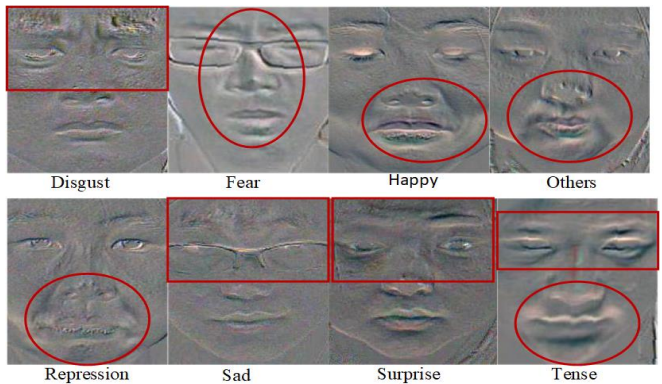

Fig.1. Dynamic responses of different micro-expressions.

In this paper, dynamic imaging and CNN-based approach (LEARNet) is cohesively used to classify the micro expressions from expression image sequences. Inspired from [16], in this paper we propose a LEARNet model, that focuses in learning salient features of micro expressions. The LEARNet is designed to process multiple feature instances through laterally connected layers. It also incorporates the accretion layer to enhance the feature learning capability of the neurons. These accretion layers culminate the hybrid feature responses by combining the previous lateral layer outcomes. Accretion layers maintain the feature map quality by preserving the non-trivial features and improve the discriminability performance of the network. Furthermore, to represent the spatio-temporal features we have elicited the dynamic images from the video sequences, which incorporate features of the both appearance and motion dynamics. These images are processed by the LEARNet. The main contributions of the paper are summarized as follows:

1) A dynamic imaging based deep network (LEARNet) is proposed to identify the emotion class for micro expressions.
2) Dynamic imaging is used to transform the video sequences into one frame instance by conserving spatio-temporal information.
3) LEARNet: Lateral Accretive Hybrid network is designed to spot the micro level involuntary changes in an expression sequence and classify them according to basic emotion classes.

The performance of proposed method is tested on four benchmark datasets: CASME-I [33], CASME-II [34], CAS(ME)^2 [35] and SMIC [36]. Experimental results and comparative analysis show that our method has gained excellent performance as compared to the existing state-of art methods. The rest of the paper has been organized as follows: Section II presents a brief overview of existing state-of-art approaches for micro expression recognition. In Section III, the proposed framework for micro expression recognition is described with visual representation. Further, experimental results and comparative analysis is discussed in Section IV. Finally, we conclude our work in Section V.

## II. LITERATURE SURVEY

Spatio-temporal features have a vital role in identifying the micro-level information change along the temporal dimension. Based on these spatio-temporal features, classifier makes distinction between the emotion classes. Polikonsky et al. [17] introduced a 3D-gradients orientation histogram-based feature descriptor to identify the micro expressions. Zhao et al. [6] used local binary pattern descriptor in three orthogonal planes (LBP-TOP) to describe the micro expressions over time by extracting the temporal and spatial features. Moreover, Pfister et al. [7] proposed a temporal interpolation model (TIM) to normalize the video frames into a definite time period. Wang et al. [8] applied the discriminant tensor subspace analysis (DTSA) model by considering the video sequences as third order tensor to regulate the motion with spatial information. Wang et al. [8] introduced an approach to extract the motion and appearance variations by using robust PCA (RPCA) [18] and local spatiotemporal directional features (LSTD) [19] respectively. Moreover, they [9-10] defined a tensor independent color space (TICS) to describe the expressions based on the color components. The video sequence was categorized into a 4D structure: first two dimensions hold the spatial information; third dimension describe the temporal variations and fourth dimension consists of the RGB color components information. Further, they transformed the RGB color components into tensor color space to enhance the independency between the color elements. Guo et al. [20] designed centralized binary pattern (CBP) descriptor in three orthogonal planes to extract the spatiotemporal features from the facial image sequences and applied the extreme learning machine (ELM) [21] to classify the micro expressions. Liu et al. [22] extracted optical flow of image sequences and identified the action units (AUs) by dividing the facial image into expressive regions. In addition, they extracted the salient features (main directional mean optical flow) from these regions and forwarded them to the SVM [23] classifier. Wang et al. [24] presented a new approach called main directional maximal difference (MDMD) to spot the discriminative features in the micro expression.

In recent time, CNN based models like VGG-16, VGG-19 [12], ResNet [15] etc. have achieved impressive results in the field of computer vision as facial expression recognition, object detection, face identification etc. Kim et al. [25] proposed a CNN based model to learn the spatio-temporal features of the micro-expressions. Patel et al. [26] used the pretrained weights of the convolution neural network (CNN) model (VGG-facenet) to capture the salient features of each frame and then apply evolutionary search to detect the disparities between the frames. However, a single instance network would restrict its ability to learn micro level features from micro expressions.

## III. PROPOSED LEARNET

In the proposed method, we adopt the concept of dynamic imaging [27] which summarizes the subtle and involuntary movement of the micro expression image sequences into one frame/image. These dynamic images are processed through the LEARNet which learns the dynamic-aware features to interpret the class of micro expression.

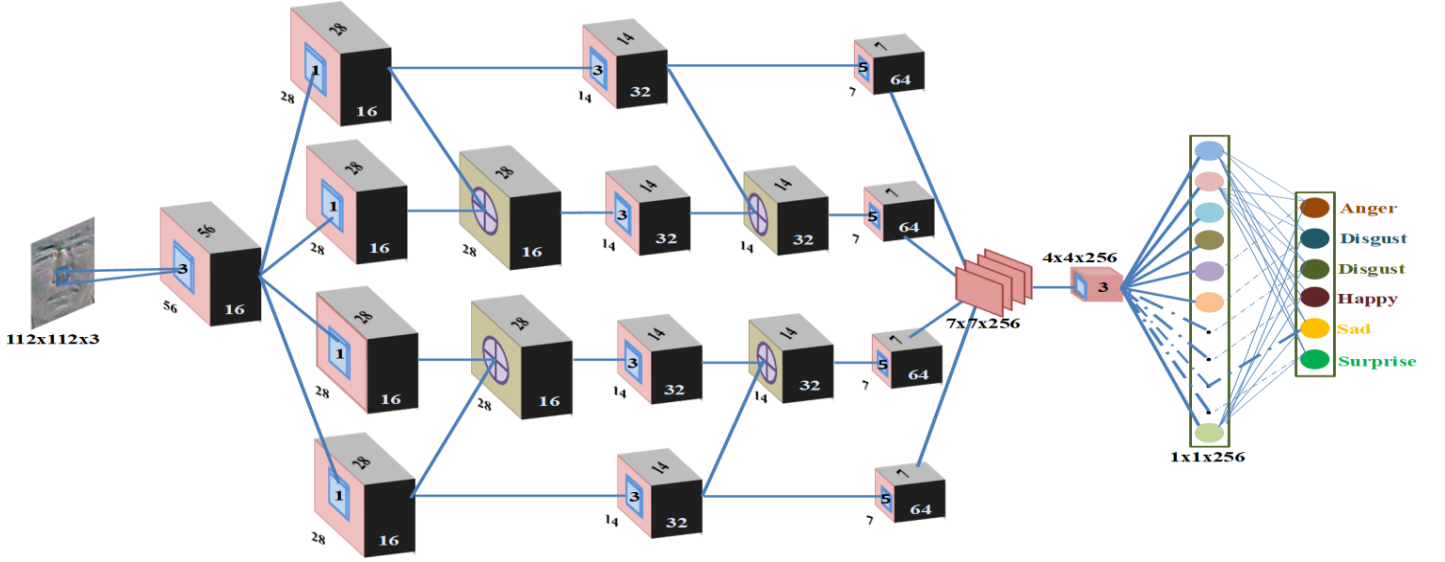

Fig. 2. Proposed LEARNet architecture

*A) Dynamic Imaging*

Micro expressions are uncontrollable instances and are generated when someone tries to control real feelings within their false emotions. As discussed in Section I, micro expressions are rapid and of short duration in nature. Thereby, they will appear only in few frames of a video. To extract these momentary changes from the video, we have used the dynamic imaging technique.

Dynamic image is a standard RGB image that holds the spatial and temporal information of a whole video sequence in a single instance. Here, we have utilized the concept of Fernando et al. [28-29] to generate exactly the same type of image. To construct dynamic image, video is referred as a ranking function by its frames $F_1, F_2, ....F_T$ i.e. $\sigma(F_t) \in R^d$ where $\sigma(F_t)$ represents the vector of RGB features, extracted from each individual frame $F_t$.

The time average ($\varphi_t$) of available feature vector is computed using Eq. (1).

$$\varphi_t = \frac{1}{t}\sum_{T=1}^{t}\sigma(F_T) \quad (1)$$

Then the ranking function calculates a score associated with time *t* by using Eq. (2).

$$\psi(t | d) = <d, \varphi_t> \quad (2)$$

where, $d \in R^d$ represents a vector which computes the frame scores in a video [30]. Higher ranks are assigned to the frames at time *l* i. e. $(l > t) \Rightarrow \psi(l | d) > \psi(t | d)$. Finally, *d* is computed by employing RankSVM [30] as formulated in Eq. (3) & Eq. (4).

$$d^* = \eta(F_1, F_2, ..., F_T; \sigma) = \arg\min(E(D)) \quad (3)$$

$$E(D) = \frac{\delta}{2}\|d\|^2 + \frac{2}{T(T-1)} \times \sum_{l>t} \max\{0, 1 - \psi(l | d) + \psi(t | d)\} \quad (4)$$

In Eq. (4), the response of two main functions are combined: first function is a quadratic regularize, mainly used in SVMs and second is a hinge-loss soft-counting function which represents that how many pairs $(l > t)$ are wrongly ranked by the rank function. However, Eq. (3) defines a function $\eta(F_1, F_2, ...F_T; \sigma)$ which transforms the video frames into a single vector **d***. Thus, d* contains sufficient information to rank all the image sequences in the video, it has cumulative information of all the frames and can be used as a video descriptor. Dynamic images of the micro expressions are demonstrated in Fig. 1. From Fig. 1, we can observe that, the resultant dynamic images (for different expression classes like disgust, fear, happy, others, repression, sad, surprise and tense) successfully preserve both uniform and non-uniform information within the single frame. The non-uniform variations play a major role in micro expression identification, highlighted by the red color boxes. These dynamic images are further fed to the proposed LEARNet for further training and inference.

*B) LEARNet Architecture*

As dynamic images of micro expressions hold minute variations within the image sequences, existing networks like VGG-16, VGG-19 [12] and ResNet [15] fail to spot these variations. These networks usually follow sequential coupling mechanism with dense depth maps. Such an approach sometimes ignore the minute features favoring more visually distinguishable features. Therefore, such networks are not quite suitable for the task of micro-expression feature learning and classification. Thus, in this paper we propose a Lateral Accretive Hybrid Network (LEARNet) to learn the minute features in the micro expression. Although, our work is inspired from the domain specific ConvNet [16] and ResNet [15] but it is substantially different in terms of the feature coupling

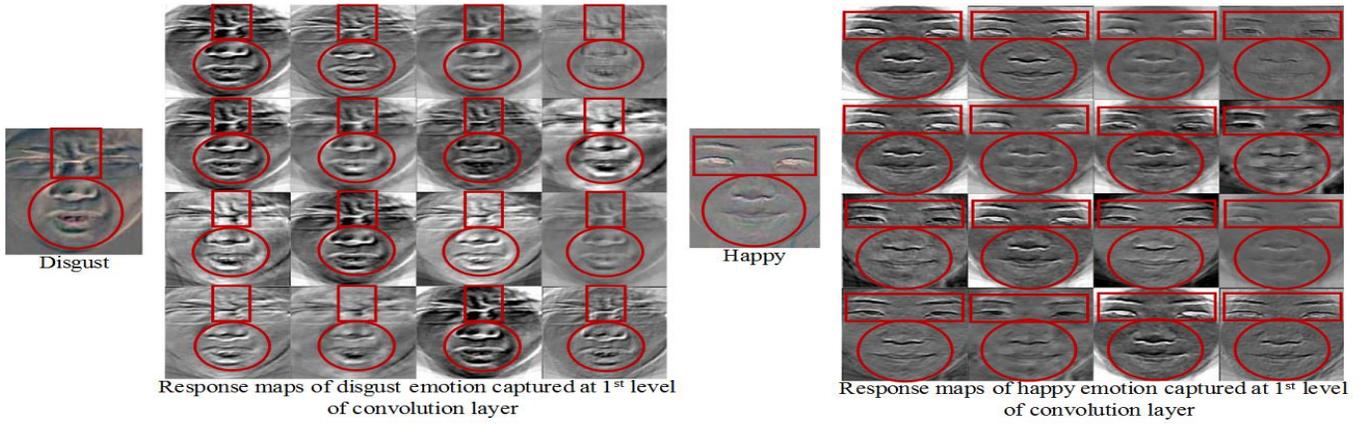

Fig. 3. Response maps of two different emotion classes a) disgust and b) happy, captured at 1st level of the convolution layer.

between the layers of the LEARNet. LEARNet begins with convolution layer which imposes 16 filters of size $3 \times 3$, which are carried out by four lateral pathways: two of them follow the sequential CNN flow and remaining two utilize the hybrid feature approach by incorporating an accretion layer to extend the network in accretive way. Accretion layer combines the hybrid responses which are generated by previous layers. These layers enhance the learnability of the neurons for minute details and maintain the essence of the feature maps. Each path consists of 16, 32 and 64 filters of size $1 \times 1$, $3 \times 3$ and $5 \times 5$ respectively. After getting feature maps from all the paths, the feature responses are concatenated and forwarded to next convolution layer with 256 filter of size $3 \times 3$ as shown in Fig. 2. Then, resultant feature maps are forwarded to fully connected layer which classifies expression image into its emotion class. The detailed explanation of architecture is given as follows:

***Convolution layer:*** Convolution layer applies convolution operation to extract the features from the input image and then feed them to the next layer. This layer contains set of neurons with learnable weights and biases. The weights of the neurons are updated according to activation map when some new feature needs to be added.

Let $F(n,n)$ ($F_M^n$) be an input image of size $M \times M$ and $f_w$ is the filter with a kernel size $k \times k$. Where $b^k$ represents the biases. The outcome of the convolutional layer is computed by Eq. (5).

$$R_M^n = f_w^k \otimes F_M^n + b^k \quad (5)$$

To calculate more precise and assorted features of the input, multiple filters $f_k$ can be used, where $k \in N$.

In LEARNet, convolution with stride 2 is utilized to reduce the size of input instead of pooling layer. Because, several layers of max pooling lose the micro level edges which are pertinent to the emotion classification. The outcome of the convolutional layer with stride 2 is computed by Eq. (6).

$$R_{M/2}^n = f_w^k \otimes F_M^{2n-1} + b^k \quad (6)$$

***Rectified linear unit***: This layer uses a monotonic function which transforms the linear input into non-linear. The activation function for ReLU is computed by Eq. (7).

$$f(\varphi) = \max(0, \varphi) \quad (7)$$

ReLU creates a sparse feature representation which makes the network more efficient. It also reduces the problem of vanishing gradient, which affects the learning process critically in earlier approaches like tanh or sigmoid. Also, ReLU helps in faster training and have cheap computational cost as compared to earlier approaches.

***Accretion Layer:*** Motivated from ResNet [15], we utilize the accretion layer in LEARNet model. Accretion layer accretes the learning capability of the network by combining the pertinent response features of the former layers by using Eq. (8).

$$A_{acc}(x) = A^1(x) + A^2(x) \quad (8)$$

Where, $A^1$ and $A^2$ are two input feature maps to the accretion layer.

***Concatenation Layer:*** Concatenation layer accumulates the multiple instances into single output instance along with 3rd-dimension. In LEARNet, this layer concatenates the output of each lateral pathway to preserve all response features of micro expressions and then forwards the same to the next layer.

***Local Response Normalization (LRN):*** LRN is used to resolve the problem of feature distribution which are diverged across the image sets: training and testing data. Moreover, LRN also enhances the strength of the network by normalizing the outcome of the previous contact layer by subtracting batch mean and dividing by the standard deviation as follows:

$$y_k \leftarrow \mu \hat{x}_k + \nu \equiv LRN_{\mu,\nu}(x_k) \quad (9)$$

$$\hat{x} = \frac{x_k - m_B}{\sqrt{d_B^2 + \epsilon}} \quad (10)$$

where, $x_k$ implies for the mini batch size $B = \{x_1, x_2, ... x_n\}$ and $\mu, \nu$ are the learnable parameters. $m_B$ and $d_B$ represent the batch mean and batch variation as calculated using Eq. [(11)-(12)].

$$m_B = \frac{1}{n} \sum_{k=1}^{n} x_k \quad (11)$$

$$d_B = \frac{1}{n} \sum_{k=1}^{n} (x_i - m_B)^2 \quad (12)$$

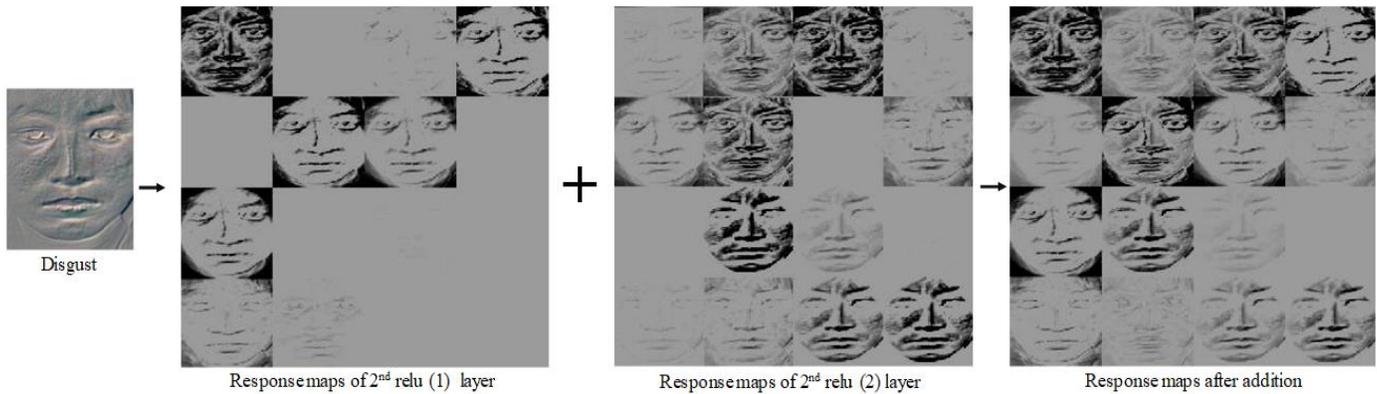

Fig. 4. Visualization of neuron responses for disgust emotion triggered by: a) Conv- 2.1 b) Conv- 2.2 and c) accretion response.
*Note*: For the better visibility inverted results are included.

**Fully Connected Layer:** In this layer neurons are fully connected to all activations in the previous layer, similar to multi-perceptron neural networks. The activation of these neurons are computed as matrix multiplication added by a bias offset. Let the input be $\alpha$ with size $m$ and $l$ be the number of neurons in the fully connected layer. The output of fully connected layer is calculated by using Eq. (13).

$$f(\alpha) = \psi(W * \alpha) \qquad (13)$$

where $\psi$ implies for the activation function and resultant matrix is $F_{m \times l}$.

**Dropout Layer:** Dropout layer was introduced by Srivastava et al. [31] to improve the sensitivity of the neurons for specific weights and resolve the problem of overfitting of the training data. The configuration details of the LEARNet network are depicted in Table I.

*C) Analysis of Proposed LEARNet*

In literature, various CNN based models e.g. VGG-16, VGG-19 [12], AlexNet [13] and ResNet [15] have been proposed to learn the salient features for classification. These CNN models generate the feature maps by utilizing the sequential coupling property with dense network layers and have shown impressive results in the field of computer vision and machine learning applications. Usually, the response feature maps of a layer are fully dependent on the previous layers. This coupling behavior of network may lose competent minute edge information thereby degrading the performance of the MER system. However, the LEARNet model captures more detailed features by using the decoupled feature map mechanism. The input is processed by four independent parallel paths. Feature maps of each path are generated by applying multiple convolution filters. Each path layer considers only inline feature maps and is independent of other paths which help in preserving the minute facial muscle change information. Thus, generated response features enhance the discriminability of the network. From Fig. 3, we can see, that the LEARNet model is able to distinguish between the emotion-class variations. We have depicted the 1st convolution layer feature maps of two emotion categories as '*disgust*' and '*happy*' in Fig. 3, respectively. Emotion-class variations are noticeably highlighted by the proposed model.

LEARNet architecture is designed in such a way that the previous feature maps of two paths follow traditional sequential CNN approach and remaining path use hybrid approach by incorporating an accretion layer. The accretion layer combines the salient features of two different layers to secure more detailed information. As shown in Fig. 4., some neurons at Conv 2.1 and Conv 2.2 have failed to capture the salient features of the expressive regions. However, accretion layer enriches the features by joining preferable feature maps of two layers. Inspired by *Inception layer* [14], we use different filter size $1 \times 1$, $3 \times 3$ and $5 \times 5$ in the proposed network to capture the high level as well as micro level information. We observe that smaller filter size is more preferable for micro expression recognition. A bigger filter size ($7 \times 7$) may skip the minute information which is quite important in micro expression. Furthermore, LEARNet model does not include max pooling layer for down sampling unlike the other CNN based models. Basically, pooling is extracting the high-level edge information by applying max function. However, pooling loses some of the pertinent features as it neglects the micro level information and extracts only high-level features by applying max function over the input data. But in micro expressions, movement of facial muscles are subtle and short lived so any loss can become a cause of inaccurate classification. LEARNet utilizes the convolution layer with stride 2 to down sample the feature maps and capture more information as compared to max pooling. Some studies in literature [32] have shown, that replacing pooling by convolution with more stride adds inter-feature dependencies. It also learns the pooling operation by parameter sharing among channels rather than fixing it. In Fig. 5, we show a comparative analysis between max pooling and convolution with stride 2. From Fig. 5, it is evident that the convolution with stride 2 outperforms the max pooling operation to improve the network learning ability for the micro expression recognition. The adaptability of the LEARNet model is visually represented in Fig. 6. From Fig. 6, it is clear, that the proposed LEARNet is able to preserve micro variations of an expression as highlighted in the red circles and rectangles. Moreover, LEARNet consists of a smaller number of depth channels and hidden layers as compared to existing networks. Thus, it requires less computation power which increases the efficiency of the MER system.

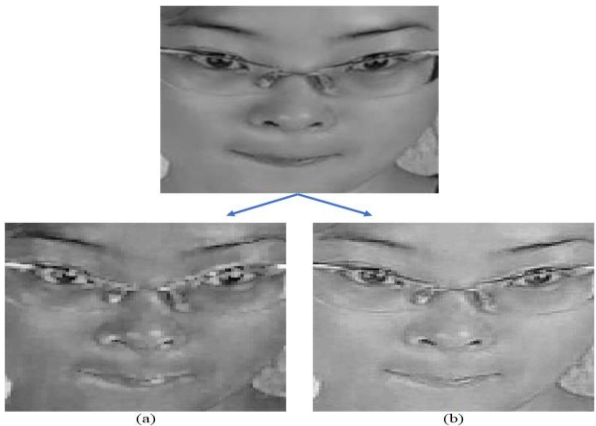

Fig. 5. The generated responses for a given sample image using a) max pooling and b) convolution with stride-2.

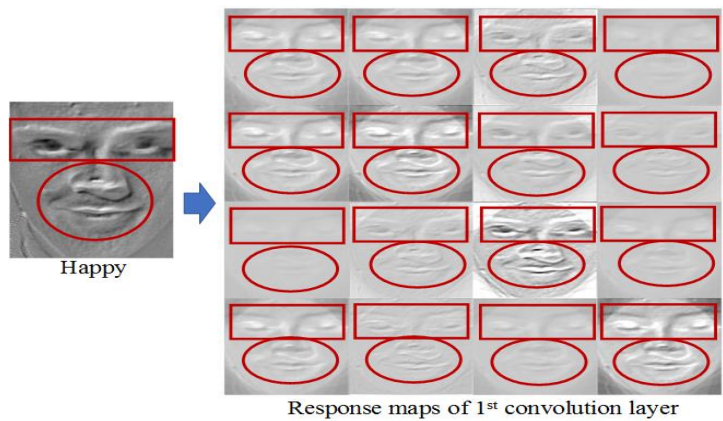

Fig. 6. The sample visual responses at $1^{st}$ Conv layer of proposed LEARNet over happy expression of SMIC dataset.

## IV. EXPERIMENTAL SETUP AND DISCUSSION

To validate the performance of the proposed approach, we have executed four experiments on benchmarked spontaneous micro expression datasets CASME-I [33], CASME-II [34], CAS(ME)^2 [35] and SMIC [36]. In the literature, two types of validation schemes are utilized to evaluate the performance of MER systems: N-fold cross validation and leave one subject out cross validation. In N-fold cross validation, the dataset is randomly divided into N folds, where N-1 folds are considered as training set and remaining assist as testing set images. Whereas, in leave one subject out cross validation, datasets are partitioned in such a way that all expressions of a particular subject at a particular iteration act as testing set and remaining data is considered as training set.

In this paper, the experiments have been conducted with N-fold cross validation. To arrange the training and testing set, we have randomly partitioned the available benchmarked datasets in offline mode with a ratio of 80:20 respectively. Furthermore, training set is also divided into training and validation set with 70:30 ratio respectively. Since, the data partition is done randomly, therefore, to make fair analysis of the outcomes, we Specifically, each dynamic image is rotated between [-45°, 45°] with the increment of 15°, enhanced by histogram equalization and then all generated images are flipped horizontally.

Moreover, to validate the effectiveness of the proposed model on our selected datasets, we have compared our results with other state-of-art approaches for MER. Due to inherent variations in dataset selection procedures and experimental setups, it is hard to justify the obtained results in comparison to existing results. To ensure a fair comparison, we have implemented the existing CNN models: VGG-16, VGG-19 [12], ResNet [15] and incorporated them into our experimental setup. In addition, we directly follow the published results of some hand-crafted methods. Furthermore, for network training we use same setup for all the experiments, we set min batch size to 25 and train up to 100 epochs. We have initialized learning rate to 1e-3 and Stochastic gradient descent (SGD) is used as optimization algorithm. All the experiments are performed on MATLAB r2018a with Titan-Xp GPU (12 GB memory) and Xeon Processor. repeat the same experiment 5 times and then average recognition rate is considered as the final accuracy. Recognition accuracy is calculated by the following Eq. (14).

$$Recog.\ Accur. = \frac{Total\ no.\ of\ correctly\ predicted\ samples}{Total\ no.\ of\ samples} \times 100 \quad (14)$$

In our experiments, we have performed data augmentation to create a larger pool of data to avoid the overfitting problem.

*1) Experiments on CASME-I Dataset*

In this experiment, the performance of the proposed LEARNet is tested over the CASME-I dataset. The CASME-I [33] comprises of 195 video clips of twenty subjects of Chinese origin. CASME-I is annotated with eight expression classes by considering three main factors: 1) action unit coding, 2) true emotion of the video and 3) self-reporting by the subject. The dataset videos have been captured with 20 fps and $150 \times 190$ face size. In this paper, we have utilized 183 video sequences: contempt-1, disgust-43, fear-2, happy-8, repression-38, sad-5, surprise-19 and tense-67, having proper class labels as mentioned in Table-II. Table III illustrates the effectiveness of the proposed LEARNet and other existing methods in terms of recognition accuracy. From Table III, it is clear that the proposed model outperforms the other existing models with comfortable margins. More specifically, it secures 44.03% and 4.23% improvement as compared to The CNN based networks: VGG-16 and ResNet, respectively. From results, it is clear that former models does not achieve high accuracy due to deep networks and large channel depths. LEARNet also obtains 16.55%, 11.76% and 6.8% more recognition rate as compared to handcrafted descriptors of LBP-TOP, MDMO with SVM classifier and LBP-TOP with ELM classifier respectively. Further, to examine the class-wise recognition accuracy (true positives and false positives), we have represented the confusion matrix in Fig. 7. From Table III and Fig. 7, it is clear that the proposed LEARNet outperforms the state-of-the-art approaches (both Deep Networks and handcrafted) on CASME-I dataset in terms of recognition accuracy.

*2) Experiments on CASME-II Dataset*

In this experiment, the performance of the proposed LEARNet is tested on the CASME-II dataset. The CASME-II dataset [34] includes 246 expression sequences captured from the twenty-six subjects with $280 \times 340$ image resolution and 200 fps device. These sequences are elicited in a well-arranged environment and proper lighting to remove the problem of illumination flickering. Among the 256 expressions, we have

TABLE I
LEARNET DETAILED CONFIGURATION

| Type | Sub-layer | Filter | Stride | Output | #Parameters (w+b) |
|---|---|---|---|---|---|
| Input | - | - | - | $112 \times 112 \times 3$ | - |
| Conv-1 | - | $3 \times 3$ | 2 | $56 \times 56 \times 16$ | 432+16 |
| Conv-2 | 2.1<br>2.2<br>2.3<br>2.4 | $1 \times 1$ | 2 | $28 \times 28 \times 16$ | $4 \times (256+16)$ |
| Add-1 | 1.1<br>1.2 | - | - | $28 \times 28 \times 16$ | - |
| Conv-3 | 3.1<br>3.1<br>3.3<br>3.4 | $3 \times 3$ | 2 | $14 \times 14 \times 32$ | $4 \times (4608+32)$ |
| Add-2 | 2.1<br>2.2 | - | - | $14 \times 14 \times 32$ | - |
| Conv-4 | 4.1<br>4.2<br>4.3<br>4.4 | $5 \times 5$ | 2 | $7 \times 7 \times 64$ | $4 \times (51200+64)$ |
| Concat | - | - | - | $7 \times 7 \times 256$ | - |
| LRN | - | - | - | $7 \times 7 \times 256$ | 256+256 |
| Conv-5 | - | $3 \times 3$ | 2 | $4 \times 4 \times 256$ | 589824+256 |
| FC | - | - | - | $1 \times 1 \times 256$ | 589824+256 |

TABLE II
THE NO. DYNAMIC IMAGES USED FOR DIFFERENT DATASETS IN THE EXPERIMENTAL RESULTS

| Emotion | CASME-I | CASME-II | CAS(ME)^2 | SMIC |
|---|---|---|---|---|
| Anger | - | - | 101 | - |
| Contempt | 1 | - | - | - |
| Disgust | 43 | 63 | 89 | 10 |
| Fear | 2 | 2 | - | 16 |
| Happy | 8 | 32 | 151 | 17 |
| Others | - | 99 | - | - |
| Sad | 5 | 8 | - | 8 |
| Surprise | 19 | 20 | - | 20 |
| Repression | 38 | 27 | - | - |
| Tension | 67 | - | - | - |
| Total | 183 | 251 | 341 | 71 |

TABLE III
RECOGNITION ACCURACY COMPARISON ON CASME-I, CASME-II AND CAS(ME)^2 DATASET

| Method | CASME-I | CASME-II | CAS(ME)^2 |
|---|---|---|---|
| LBP-TOP-SVM* [6] | 64.07 | 57.16 | - |
| LBP-TOP-ELM *[20] | 73.82 | - | - |
| MDMO-SVM* [22] | 68.86 | 67.37 | - |
| CNN-LSTN* [25] | - | 60.98 | - |
| VGG-16 [12] | 36.59 | 44.29 | 44.29 |
| VGG-19 [12] | 36.59 | 44.29 | 44.28 |
| ResNet [15] | 76.39 | 74.49 | 74.48 |
| **LEARNet** | **80.62** | **76.57** | **76.33** |

* This result is from the corresponding original paper

TABLE IV
RECOGNITION ACCURACY COMPARISON ON SMIC DATASET

| Method | 5-Class | 2-Class |
|---|---|---|
| LBP-TOP-SVM* [6] | 71.40 | - |
| MDMO-SVM *[22] | 80.00 | - |
| VGG-16 [12] | 36.59 | 51.53 |
| VGG-19 [12] | 36.59 | 51.53 |
| ResNet [15] | 71.36 | 88.27 |
| **LEARNet** | **81.60** | **91.09** |

* This result is from the corresponding original paper

considered 251 videos having seven class labels as: disgust-63, fear-2, happy-32, others-99, repression-27, sad-8 and surprise-20. Table III, shows the effectiveness of the proposed LEARNet network over exiting state-of-art methods in terms of recognition accuracy. From Table III, it is clear that the proposed network obtains higher recognition accuracy as compared to other approaches. It achieves performance gain of 32.28%, 15.59% and 2.08% over VGG-19, CNN-LSTN and ResNet respectively. It also achieves 19.41% and 9.2% more accuracy in comparison to traditional handcrafted techniques LBP-TOP and MDMO respectively. Detailed results calculated over CASME-II are illustrated in Fig. 8. Both Table III and Fig.8 validate the achievement of the proposed LEARNet as compared to other state-of-the-art (both deep networks and handcrafted) approaches.

*3) Experiments on CASME^2 Dataset*
The CAS(ME)^2 dataset [35] included 357 video sequences of 22 subjects. These videos have been captured by a camera with 500 ms. This dataset contains both spontaneous macro (300) and micro (57) expressions. Image sequences are categorized into three emotion classes: anger, happy and disgust, by extracting more than 600 AUs. Furthermore, subject's self-reviews are also considered to improve the reliability of the expression categories. In our experimental setup, we have selected total 341 image sequences: anger-101, happy-151 and disgust-89 of micro expressions. Table III demonstrates the improved accuracy results obtained by proposed LEARNet as compared to existing approaches. Specifically, the proposed method has obtained 32.04%, 1.85% more accuracy as compared to VGG-19 and ResNet respectively. Class-wise accuracy results for CAS(ME)^2 are presented in Fig. 9. Results in Table III and Fig. 9 prove that the proposed LEARNet outperforms the other existing state-of-the-art approaches (both deep networks and handcrafted).

*4) Experiments on SMIC Dataset*
We conduct our final experiment on the SMIC dataset. The SMIC dataset [36] consists of 77 spontaneous micro expression video recording of six subjects. These videos have been recorded with 100 fps and 190× 230 resolution. The captured video sequences are arranged in two formats: one categorized micro expression classes namely negative/positive and another contained five classes as: disgust, fear, happy, sad and surprise. In our experimental setup, we have included 35 videos of negative-18 and positive-17 micro expression classes and named the dataset as SMIC-I. On the other hand, 71 videos are included in five classes: disgust-10, fear-16, happy-17, sad-8 and surprise-20 and names as SMIC-II. Table-III denotes the superior performance of proposed model over other approaches. Particularly, the proposed model gains 45.01% and 10.24% extra recognition accuracy for 5 class expressions as compared to CNN based approaches VGG-19 and Resnet models respectively. Further, it yields 5.03% and 10.2%, better accuracy as compared to the handcrafted techniques of LBP-Top and MDMO respectively. In another dataset SMIC-II, our approach yields 39.56% and 2.82% better accuracy rate as compared to VGG-19 and ResNet respectively. The confusion matrix for the testing results over SMIC dataset is shown in Fig.

|  | \multicolumn{8}{c}{TRUE LABEL} |
|---|---|---|---|---|---|---|---|---|
|  | Con | Dis | Fea | Hap | Rep | Sad | Sur | Ten |
| Con | 0.60 | 0.17 | 0.01 | 0.04 | 0.06 | 0.02 | 0.05 | 0.05 |
| Dis | 0.02 | 0.88 | 0.04 | 0.02 | 0.01 | 0.01 | 0.01 | 0.01 |
| Fea | 0.07 | 0.09 | 0.73 | 0 | 0.01 | 0.06 | 0.01 | 0.03 |
| Hap | 0.05 | 0.02 | 0 | 0.89 | 0.02 | 0 | 0.02 | 0 |
| Rep | 0.01 | 0.02 | 0.03 | 0 | 0.87 | 0.02 | 0.01 | 0.04 |
| Sad | 0.04 | 0.02 | 0.04 | 0.02 | 0.08 | 0.71 | 0.03 | 0.06 |
| Sur | 0.01 | 0.01 | 0.02 | 0.04 | 0.01 | 0.01 | 0.88 | 0.02 |
| Ten | 0.01 | 0.01 | 0.02 | 0 | 0.03 | 0.04 | 0 | 0.89 |

Fig. 7. Confusion matrix of LEARNet for 8-class expression classification in CASME-I dataset. *(Con: Contempt, Dis: Disgust, Fea: Fear, Hap: Happy, Rep: Repression, Sad, Sur: Surprise, Ten: Tension)*

|  | Dis | Fea | Hap | Oth | Rep | Sad | Sur |
|---|---|---|---|---|---|---|---|
| Dis | 0.87 | 0.04 | 0.02 | 0.01 | 0.03 | 0 | 0.03 |
| Fea | 0.03 | 0.59 | 0.05 | 0.12 | 0.09 | 0.09 | 0.03 |
| Hap | 0.09 | 0 | 0.77 | 0.07 | 0.05 | 0 | 0.02 |
| Oth | 0.03 | 0 | 0 | 0.92 | 0.01 | 0.01 | 0.03 |
| Rep | 0 | 0.04 | 0.04 | 0.06 | 0.83 | 0.03 | 0.02 |
| Sad | 0.04 | 0.09 | 0.08 | 0.06 | 0 | 0.67 | 0.06 |
| Sur | 0.11 | 0 | 0.09 | 0.02 | 0.05 | 0.02 | 0.71 |

Fig. 8. Confusion matrix of LEARNet for 7-class expression classification in CASME-II dataset. *(Dis: Disgust, Fea: Fear, Hap: Happy, Oth: Other, Rep: Repression, Sad and Sur: Surprise).*

|  | ANGER | DISGUST | HAPPY |
|---|---|---|---|
| ANGER | 0.78 | 0.17 | 0.05 |
| DISGUST | 0.14 | 0.70 | 0.16 |
| HAPPY | 0.11 | 0.08 | 0.81 |

Fig. 9. Confusion matrix of LEARNet for 3-class expression classification in CAS(ME)^2 dataset.

|  | Dis | Fea | Hap | Sad | Sur |
|---|---|---|---|---|---|
| Dis | 0.89 | 0.01 | 0.02 | 0.06 | 0.02 |
| Fea | 0.04 | 0.73 | 0.03 | 0.15 | 0.05 |
| Hap | 0.06 | 0 | 0.87 | 0.01 | 0.08 |
| Sad | 0 | 0.17 | 0.08 | 0.71 | 0.04 |
| Sur | 0.02 | 0.04 | 0.06 | 0 | 0.88 |

Fig. 10. Confusion matrix of LEARNet for 5-class expression classification in SMIC-I dataset. *(Dis: Disgust, Fea: Fear, Hap: Happy, Sad and Sur: Suprise)*

TABLE V
COMPUTATIONAL COMPLEXITY ANALYSIS OF LEARNET AND EXISTING NETWORKS

| Network | # Layers | # Parameters (in millions) |
|---|---|---|
| VGG-16 [12] | 16 | 138 |
| VGG-19 [12] | 19 | 144 |
| GoogleNet [13] | 22 | 4 |
| ResNet [15] | 34 | 11 |
| **LEARNet** | **14** | **1.4** |

10. From the Table IV and Fig. 10, it is clear that the proposed LEARNet yields better performance as compared to other state-of-the-art approaches (both deep networks and handcrafted).

*5) Qualitive Analysis*

The qualitive visual analysis of the existing and proposed model is depicted in Fig. 11. This figure contains the visual representation of different emotion classes as *tension, happy, anger and surprise* from all four datasets: CASME-1, CASME-2, CASME^2 and SMIC respectively. In Fig. 11, we have depicted two most prominent visual responses generated by intermediate hidden layers. Expressive regions like eyes, nose and mouth which play a significant role in defining disparities between emotion class are highlighted with red boxes. It is clear from Fig. 11, that the dynamic representation of the image sequences significantly assist in preserving the minute variations in different expressive regions in the facial image. For example, in tension: glabella, eyes; *in happiness: eyes, mouth; in anger: eyes, eyebrows and in surprise: forehead, eyes, lips* regions give maximum affective response for related micro-expressions. Furthermore, the LEARNet captures micro level variational features for more accurate emotion classification. From Fig. 11, we can conclude that LEARNet has preserved more relevant feature responses to outperform the existing CNN based networks VGG-19 and ResNet for almost all emotion classes.

*6) Computational Complexity*

This section provides the comparative analysis of the computational complexity between the existing and proposed network. The total number of parameters involved in each network are tabulated in Table V. The proposed LEARNet has only 1.4 million learnable parameters which are very less as compare to other existing benchmark models like: VGG-16: 138M, VGG-19: 144M, GoogleNet: 4M and ResNet: 11M. Moreover, LEARNet architecture has fewer depth channels and hidden layers as compare to former methods. Particularly, LEARNet comprises of 14 layers. In comparison to that, VGG-16, VGG-19, GoogleNet and ResNet consists of 16, 19, 22 and 34 layers respectively.

V. CONCLUSION

This paper presents a dynamic image-based network named LEARNet for micro expression recognition. Firstly, we have generated dynamic images from micro expression sequence which captures the facial movements in one frame. Furthermore, dynamic images are processed through the LEARNet architecture for training and inference. The proposed architecture adopts hybrid and decoupled feature learning mechanism to learn the salient features from the expressive regions captured in the past layers. The feature learning is carried out by four independent paths which extract features in their own way. Out of the four paths, two paths use hybrid response map approach in which feature map of other paths will be added to extract salient features. This architecture leads to effective learning of salient features and unveils the hidden appearance information on face more accurately. Furthermore, the LEARNet uses different sized filters i.e. 1x1, 3x3 and 5x5, which enhance the capability of network by extracting micro and high-level features. Moreover, proposed network includes the accretion layer to merge the features of two response maps that allow to expose pertinent features robustly. The proposed network has a small number of parameters that reduce the training and testing time of the LEARNet system. The effectiveness of system evaluated on the benchmark dataset CASME-I, CASME-II, CAS(ME)^2 and SMIC. It is evident from experimental results that LEARNet has achieved better accuracy rates as compared to the state-of-the-art approaches. Specifically, proposed LEARNet has achieved 43.83%, 32.10%, 32.20% and 39.56% more recognition rates as compare to VGG-19. Similarly, LEARNet has obtained 4.03%, 1.90%, 1.79% and 2.82% more accuracy as compared to ResNet network on four benchmark datasets.

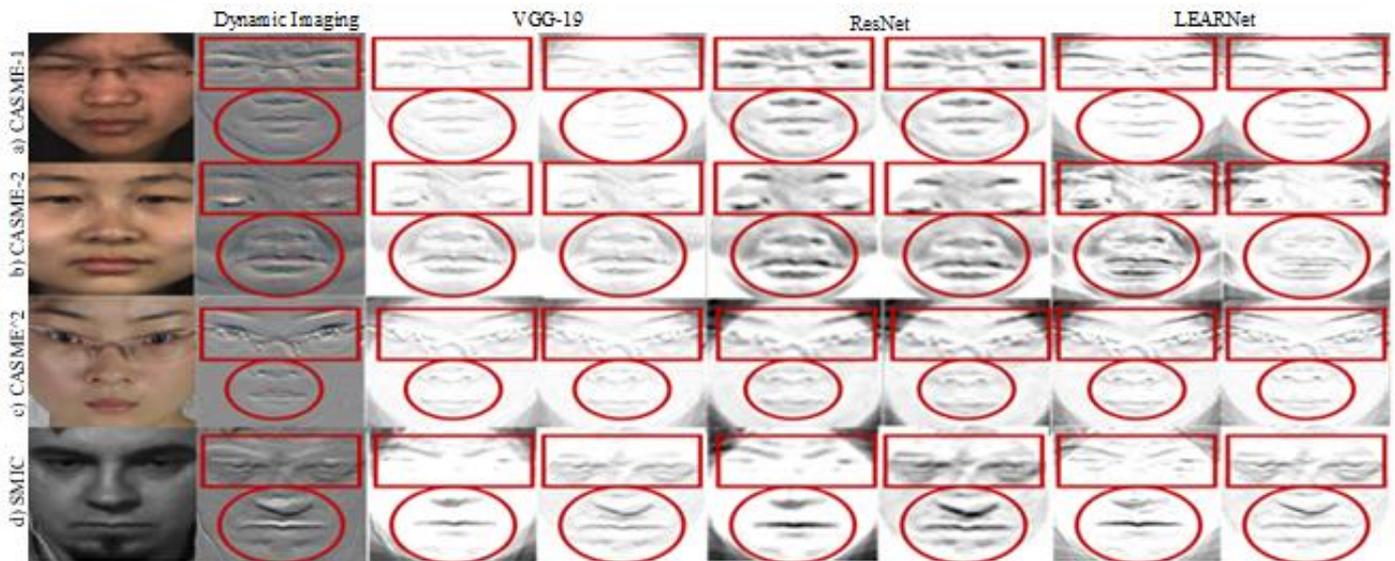

Fig.11. Visual comparison of existing model and LEARNEet over different expression of four datasets a) CASME-I: Tension b) CASME-II: Happy c) CAS(ME)^2: Anger and d) SMIC: surprise.

## ACKNOWLEDGMENTS

This work was supported by the Science and Engineering Research Board (under the Department of Science and Technology, Govt. of India) project #SERB/F/9507/2017. The author would like to thank our Vision Intelligence lab group for their valuable support. We are also thankful to NVIDIA for providing TITAN XP GPU grant.

## REFERENCES

[1] P. Ekman, W. V. Friesen, "Nonverbal leakage and clues to deception," *Psychiatry*, vol. 32, pp.88-106, 1969.

[2] M. Frank, M. Herbasz, K. Sinuk, A. Keller and C. Nolan, "I see how you feel: Training laypeople and professionals to recognize fleeting emotions," In *the Annual Meeting of the International Communication Association. Sheraton New York, New York City.*, 2009

[3] P. Ekman, "Facial expression and emotion," *American psychologist*, vol. 48, no.4, pp. 384, 1993.

[4] P. Ekman, and W. V. Friesen, "Facial action coding system," 1977.

[5] P. Ekman, "Micro expression training tool (METT)," *CD-ROM. Oakland*, 2003.

[6] G. Zhao, and M. Pietikäinen, "Dynamic texture recognition using local binary patterns with an application to facial expressions," *IEEE Trans. Pattern Analy. Mach. Intel*., vol 29, no. 6, pp.915-928, 2007.

[7] T. Pfister, X. Li, G. Zhao and M. Pietikäinen, "Recognising spontaneous facial micro-expressions," in *Proc. IEEE Int. Conf. Comput. Vis.*, pp. 1449-1456,2011.

[8] S. J. Wang, H. L. Chen, W. J. Yan, Y. H. Chen and X. Fu, "Face recognition and micro-expression recognition based on discriminant tensor subspace analysis plus extreme learning machine," *Neural Process. letters*, vol. 39, no. 1, pp.25-43, 2014.

[9] S. J. Wang, W. J. Yan, G. Zhao, X. Fuand C. G. Zhou, "Micro-expression recognition using robust principal component analysis and local spatiotemporal directional features," in *Proc. Springer Conf. Comput. Vis.,* pp. 325-338, 2014.

[10] S. J. Wang, W. J. Yan, X. Li, G. Zhao and X. Fu, "Micro-expression recognition using dynamic textures on tensor independent color space," in *Proc. IEEE Conf. Pattern Recognit.,* pp. 4678-4683, 2014.

[11] S. J. Wang, W. J. Yan, X. Li, G. Zhao, C. G. Zhou, X. Fu, M. Yang and J. Tao, "Micro-expression recognition using color spaces," *IEEE Trans. Image Process.*, vol. 24, no. 12, pp.6034-6047, 2015.

[12] K. Simonyanand A. Zisserman, "Very deep convolutional networks for large-scale image recognition,". *arXiv preprint arXiv:1409.1556,* 2014.

[13] A. Krizhevsky, I. Sutskever and G. E. Hinton, "Imagenet classification with deep convolutional neural networks," In *Neural Information Process. Systems,* pp. 1097-1105, 2012.

[14] C. Szegedy, W. Liu, Y. Jia, P. Sermanet, S. Reed, D. Anguelov, D. Erhan, V. Vanhoucke and A. Rabinovich, "Going deeper with convolutions," In *Proc. IEEE Conf. Comput. Vis. Pattern Recognit.,* pp. 1-9, 2015.

[15] K. He, X. Zhang, S. Ren and J. Sun, "Deep residual learning for image recognition," In *Proc. IEEE conf. Comput. Vis. Pattern Recognit.*, pp. 770-778, 2016.

[16] J. Tompson, K. Schlachter, P. Sprechmann and K. Perlin, "Accelerating eulerian fluid simulation with convolutional networks," *arXiv preprint arXiv:1607.03597,* 2016.

[17] S. Polikovsky, Y. Kamedaand Y. Ohta, "Facial micro-expressions recognition using high speed camera and 3D-gradient descriptor," pp. 16-16, 2009.

[18] J. Wright, A. Ganesh, S. Rao, Y. Peng and Y. Ma, "Robust principal component analysis: Exact recovery of corrupted low-rank matrices via convex optimization," In *Neural Information Process. Systems*, pp. 2080-2088, 2009.

[19] G. Zhao and M. Pietikäinen, "Visual speaker identification with spatiotemporal directional features," In *Int. Conf. Springer Image Anal. Recognit.*, pp. 1-10, 2013

[20] Y. Guo, C. Xue, Y. Wang and M. Yu, "Micro-expression recognition based on CBP-TOP feature with ELM," *Optik-International Journal for Light and Electron Optics*, vol. 126, no. 23, pp.4446-4451, 2015.

[21] G. B. Huang, Q. Y. Zhuand C. K. Siew, "Extreme learning machine: theory and applications," *Neuro Computing*, vol. 70, no. 1-3, pp.489-501, 2006.

[22] Y. J. Liu, Y.K. Zhang, W. J. Yan, S.J. Wang, G. Zhaoand X. Fu, "A main directional mean optical flow feature for spontaneous micro-expression recognition," *IEEE Trans. Affect. Comput.*, vol. 7, no. 4, pp. 299-310, 2016.

[23] M. A. Hearst, S. T. Dumais, E. Osuna, J. Plattand B. Scholkopf, "Support vector machines," *IEEE Intelligent Systems and their applications*, vol. 13, no. 4, pp.18-28, 1998.

[24] S. J. Wang, S. Wu, X. Qian, J. Liand X. Fu, "A main directional maximal difference analysis for spotting facial movements from long-term videos," *Neuro Computing*, vol. *230*, pp.382-389, 2017.

[25] D. H. Kim, W. J. Baddarand Y. M. Ro, "Micro-expression recognition with expression-state constrained spatio-temporal feature representations," In *Proc. ACM Multimedia,* pp. 382-386, 2016.

[26] D. Patel, X. Hong and G. Zhao, "Selective deep features for micro-expression recognition," In *Proc. Int. Conf. Pattern Recognit.* pp. 2258-2263, 2016.

[27] H. Bilen, B. Fernando, E. Gavves, A. Vedaldiand S. Gould, "Dynamic image networks for action recognition," In *Proc. IEEE Int. Conf. Comput. Vis. Pattern Recognit.* pp. 3034-3042, 2016.

[28] B. Fernando, P. Anderson, M. Hutter, S. Gould, "Discriminative hierarchical rank pooling for activity recognition," In *Proc. Int. IEEE Conf. Comput. Vis. Pattern Recognit.,* pp. 1924-1932, 2016.

[29] B. Fernando, E. Gavves, J. M. Oramas, A. Ghodrati and T. Tuytelaars, T., 2015. Modeling video evolution for action recognition," In *Proc. Int. IEEE Conf. Comput. Vis. Pattern Recognit.,* pp. 5378-5387, 2015.

[30] A. J. Smola and B. Scholkopf. A tutorial on support vector ¨ regression. Statistics and computing, 14:199–222, 2004.

[31] N. Srivastava, G. Hinton, A. Krizhevsky, I. Sutskeverand R. Salakhutdinov, "Dropout: a simple way to prevent neural networks from overfitting," *The Journal of Machine Learning Research*, vol 15, no. 1, pp.1929-1958, 2014.

[32] J. T. Springenberg, A. Dosovitskiy, T. Brox and M. Riedmiller, "Striving for simplicity: The all convolutional net," *arXiv preprint arXiv:1412.6806, 2014*.

[33] W. J. Yan, Q. Wu, Y. J. Liu, S. J. Wang and X. Fu, "CASME database: a dataset of spontaneous micro-expressions collected from neutralized faces," In *Proc. IEEE Int. Conf. Auto. Face and Gesture Recognit.* pp. 1-7, 2013.

[34] W. J. Yan, X. Li, S. J. Wang, G. Zhao, Y. J. Liu, Y. H. Chenand X. Fu, "CASME II: An improved spontaneous micro-expression database and the baseline evaluation," *PloS one*, vol. *9, no.* 1, p.e 86041, 2014.

[35] F. Qu, S. J. Wang, W.J. Yan, H. Li, S. Wu and X. Fu, "CAS (ME)^ 2: a database for spontaneous macro-expression and micro-expression spotting and recognition," *IEEE Trans. Affect. Comput., 2017*.

[36] X. Li, T. Pfister, X. Huang, G. Zhao and M. Pietikäinen, "A spontaneous micro-expression database: Inducement, collection and baseline," In *Proc. IEEE Int. Conf. Auto. Face and Gesture Recognit.* pp. 1-6, 2013.

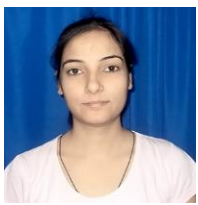

**Monu Verma** received her B. Tech degree in Computer Science and Engineering from Govt. Engineering College Bikaner, India, in 2013. She received her M. tech degree in 2016 from National Institute of Technology Jalandhar, India. She is currently pursuing her Ph.D. with the Department of Computer Science at MNIT, Jaipur, India. Her current research interests include facial expression recognition, depression analysis, and micro expression analysis.

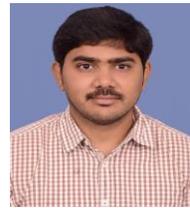

**Santosh Kumar Vipparthi** was born in Komarada, Andhra Pradesh, India in 1985. He received his B.E. degree in Electrical and Electronics Engineering from Andhra University, Andhra Pradesh, India, in 2007. Further, he received his M. Tech. and Ph. D. in Systems Engineering from IIT BHU, Varanasi, India, in 2010 and 2014 respectively. Currently, he is working as an assistant professor in the department of computer science and engineering, MNIT, Jaipur, Rajasthan, India. His major fields of interest are micro/ macro expression recognition, face recognition, change detection, design and development of feature descriptors for natural / medical image / video analysis and retrieval, real-time system analysis.

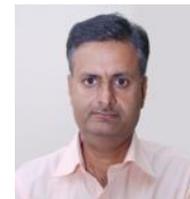

**Girdhari Singh** received the B.E. degree in Computer Engineering from Amravati University, Maharastra India, in 1990. Afterwards, he received his MS in Software Engineering from BITS Pilani, India in 1996. Further, He received his Ph.D. in Computer Engineering from MNIT, Jaipur, India in 2009. Currently, he is working as an associate professor in the department of computer science and engineering, MNIT, Jaipur, Rajasthan, India. His major fields of research are software engineering, intelligent systems image processing and machine learning.

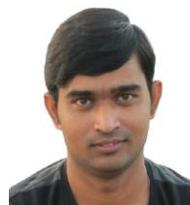

**Subrahmanyam Murala** received the B.E. degree in electrical and electronics engineering from Andhra University, Visakhapatnam, India, in 2007, and the M.Tech. and Ph.D. degrees from the Department of Electrical Engineering, IIT Roorkee, Roorkee, India, in 2009 and 2012, respectively. He was a Post-Doctoral Researcher with the Department of Electrical and Computer Engineering, University of Windsor, Windsor, ON, Canada, from 2012 to 2014. He is currently an Assistant Professor with the Department of Electrical Engineering, IIT Ropar, Rupnagar, India. His major fields of interests are computer vision, medical image processing, image retrieval, and object tracking.